\documentclass[english]{article}
\usepackage{graphicx}
\usepackage{tabularx}
\usepackage{soul}
\usepackage{epstopdf}
\usepackage[utf8]{inputenc}
\usepackage{hyperref}
\usepackage{xstring}
\usepackage{enumitem}
\usepackage{color}
\usepackage{authblk}
\usepackage{tikz}
\usepackage{pbox}
\usepackage{comment}
\usepackage{dblfloatfix}
\usepackage{multirow}

\usepackage{geometry}
\begin{document}

\title{Example-Based Machine Translation\\from Text to a Hierarchical Representation of Sign Language}

\author{
    \'Elise Bertin-Lemée\textsuperscript{1}, Annelies Braffort\textsuperscript{2}, Camille Challant\textsuperscript{2}, Claire Danet\textsuperscript{2}, Michael Filhol\textsuperscript{2} \\
    \textsuperscript{1}SYSTRAN, 5 rue Feydeau, Paris, France, elise.bertinlemee@systrangroup.com \\
    \textsuperscript{2}Universit\'e Paris-Saclay, CNRS, LISN, Orsay, France, \\
    \{annelies.braffort, camille.challant, claire.danet, michael.filhol\}@lisn.upsaclay.fr
}
\maketitle

\begin{abstract}
This article presents an original method for Text-to-Sign Translation.
It compensates data scarcity using a domain-specific parallel corpus of alignments between text and hierarchical formal descriptions of Sign Language videos in AZee.
Based on the detection of similarities present in the source text, the proposed algorithm recursively exploits matches and substitutions of aligned segments to build multiple candidate translations for a novel statement.
This helps preserving Sign Language structures as much as possible before falling back on literal translations too quickly, in a generative way.
The resulting translations are in the form of AZee expressions, designed to be used as input to avatar synthesis systems.
We present a test set tailored to showcase its potential for expressiveness and generation of idiomatic target language, and observed limitations.
This work finally opens prospects on how to evaluate translation and linguistic aspects, such as accuracy and grammatical fluency.
\end{abstract}

\section{Introduction}

Rosetta\footnote{\url{https://rosettaccess.fr/index.php/home-page-english/}} is a French project that aimed to study accessibility solutions for audiovisual content. One of the experiments consisted in designing an automatic translation system from text to Sign Language (SL) displayed through animation of a virtual signer. 

The three main contributions concerning SL in this project were the constitution of Rosetta-LSF \cite{LREC-2022-corpusRosetta}, an aligned corpus of text and SL captured using a mocap system, a transflation system from text to AZee (a representation of SL content), and a system allowing to generate virtual signer animations from AZee input.

This article describes the second contribution: the system of translation from text to AZee. 
After an overview of the issues and recent works in the field, we explain our method and design choices and describe the implementation of the translation system. Finally, we give preliminary results and discuss the questions raised for evaluation.

\section{Text-to-Sign translation}

The automatic translation of content from a spoken language into a sign language is a fairly recent and still largely unexplored research topic. Here we are interested in the translation of text as the source format, in our case in French, and video or 3D animation as the target format, in our case French Sign Language (LSF).

In this section, we look at the main challenges encountered with text-to-sign translation.

\subsection{Need for bilingual corpora}

Machine translation (MT) was first developed for spoken languages in their written form using bilingual dictionaries and rule-based systems, that were not easy to develop and maintain. Access to parallel corpora of aligned examples has led to the rise of data-driven approaches, such as Statistical Machine Translation (SMT) that used the frequencies of translation pairs containing source--target pairings of words or phrases. In the current dominant approach, Neural Machine Translation (NMT), which is also data-driven, the source text is encoded into an intermediate representation in the form of numerical vectors to be decoded as a target text. Although the representation is not directly open to interpretation, the practical results largely prevail over former strategies. These methods designed for spoken languages rely on the availability of large volumes of parallel data (of the order of several million sentences), Sign languages being considered as low resource in this respect. 

Example-based MT (EBMT) is another data-driven approach based on analogy \cite{nagao:ahi84}. It uses a bilingual corpus that contains texts and their translations. Given a sentence to translate, sentences from this corpus are selected that contain similar sub-sentential components. The similar sentences are then used to translate the sub-sentential components of the original sentence into the target language, and these phrases are put together to form a complete translation. Although the larger the corpus, the better the results will be, this approach can be implemented on smaller corpora and thus may be considered in the case of Sign Language Machine Translation (SLMT).

\subsection{Need for an intermediate representation}

One of the major differences between SLMT and written MT is the difference in channel. Written languages are input to MT systems as streams of discrete tokens---words separated by blanks---whereas SL does not have a written form. SLs are to be considered as face-to-face oral languages. Moreover, they are able to convey simultaneous information by the way of a number of articulators, such as the two hands and arms, but also the torso, shoulders, head, gaze and facial expressions (including a number of facial component movements).

As SL has no written form, many approaches proceed in two steps: a first step transforms the SL content into an intermediate representation, and a second step uses this representation as the input of a synthesis system to control the animation of an avatar in order to display the content in SL.

After a first generation of studies based mainly on the rule-based approach \cite{veale1998challenges,zhao2000machine,marshall-04}), a few ones have investigated Example-Based Machine Translation \cite{morrissey2005example}. They have sometimes be combined with statistical approaches, such as \cite{demartino2017signing} who have developed a system that automatically translates Brazilian Portuguese text to Brazilian SL (LIBRAS) by combining SMT with EBMT in case of unseen texts or ambiguous terms dependent on the context and frequency of occurrence in previous translations. To our knowledge, these projects have not led to any follow-up, nor to consumer applications.

The vast majority of projects using an intermediate representation of SL, including the latest ones \cite{gomez2021}, use sequences of glosses, each gloss\footnote{A gloss is a text label, generally a single word, reflecting the meaning of the sign it stands for.} standing for a so-called lexical unit generally restricted to manual activity.
The translation systems then deal with a sequence of tokens and as such, meet the requirements for the approaches designed for sequences.
With this kind of representation though, it is very difficult if not impossible to the handle common SL phenomena like non-manual activity, spatial relations, depicting structures, or the rhythm of the signing production. This results in low quality animations, incomplete if not incomprehensible, and therefore unacceptable by the Deaf community. For this reason, it seems important to consider a richer intermediate representation than mere concatenations of glosses.

Note that in some recent neural-based approaches \cite{stoll2020text2sign}, the use of an intermediate representation is not present. This neural-based approach generates directly photo-realistic continuous sign videos from text inputs. This is still very experimental, and does not offer anonymisation, unlike virtual signer output. Moreover, an avatar's appearance can be suited to different use cases or audiences. For these reasons, avatar-based approaches seem more flexible and appealing.

This work therefore chose to explore EBMT for translation to SL, given that we do not have a large bilingual corpus and that we consider an intermediate representation for SL more appropriate.

\section{Method}

In view of an EBMT approach as explained above, we have created \emph{Rosetta-LSF} \cite{LREC-2022-corpusRosetta} and selected the intermediate representation \emph{AZee} to represent the SL utterances, which we explain in this section.

\subsection{EBMT-type approach}

\label{substitutions}
As explained above, EBMT is a translation mechanism based on analogy from examples.
This means that we can compensate for a missing example by finding one close enough, and working from it to replace what is different.
For example, to translate ``\textit{la présidente parle nerveusement}'' (the president is speaking nervously) when the example is not in the data base of examples, we can hope to work from the translation of ``\textit{le ministre de l'écologie parle nerveusement}'' (the minister of the environment is speaking nervously), with a substitution.

In such candidate segment henceforth, we will call ``anti-matches'' the parts that do not match the query in the segment that otherwise does, and ``corrections'' the respective text parts that would have been a match.
For example, ``\textit{présidente}'' (president) is the anti-match above, and ``\textit{ministre de l'écologie}'' (minister of the environment) its correction.

A hypothesis is that if we find the portions corresponding to each anti-match in the aligned translation, we can attempt to replace them with translations of their corrections.

Our aim is therefore to produce a translation of the source written text into the chosen intermediate representation that reflects the target signed language, AZee.

\subsection{AZee} \label{AZee}
AZee is a formal approach to SL discourse representation \cite{filholNonManualFeaturesRight2014}. It allows to define \emph{production rules} that associate forms to articulate (e.g.\ begin eyebrow raise before $X$) to identified meaning (semantic operations, e.g.\ expression of $Y$ with doubt). By combining them, one can build hierarchically structured \emph{discourse expressions} representing full discourse utterances, determining the forms to produce while exposing the meaning.

For example, consider the four productions rules below:
\begin{itemize}
  \item \texttt{info-about}(\emph{topic}, \emph{info}): \emph{info}, which is focused, is given about a \emph{topic};
  \item \texttt{nerveusement}(\emph{sig}): \emph{sig} in a nervous, stressed out way;
  \item \texttt{président}: president;
  \item \texttt{parler}: speak.
\end{itemize}

These can be combined in the expression below, which not only creates a semantic combination interpretable as ``the president is speaking nervously'', but also produces, through recursive application of each rule's forms, the resulting overall signed form with that meaning. A corpus of 120 such expressions has been published by Challant \& Filhol \cite{LREC-2022-corpusAZee}.

\begin{minipage}[t]{1\columnwidth}
\begin{verbatim}
:info-about
  'topic
  :président
  'info
  :nerveusement
    'sig
    :parler
\end{verbatim}
\end{minipage}
\medskip

Since it represents the articulations necessary to convey that meaning, it can be used as the output of our translation system, a lot easier than attempting to generate video frames directly.
Of course this requires to append an animation system to the pipeline, able to render AZee input to SL video. But this is outside the scope of this paper, and assumed for now.

AZee discourse expressions are hierarchical, each nested expression covering a sub-part of the discourse.
So unlike a linear stream like text or video where segments are typically specified with start and length, identifying an AZee ``segment'' can be done through identification of a single node in the expression. This node is the root of a sub-tree (or a leaf) which covers a time segment in the video (the SL capture modelled with the expression, or indeed any avatar animation rendered from the expression).

\subsection{Corpus}
As explained above, we needed a bank of alignments between French text segments and AZee expression nodes.
For this purpose, we worked on a parallel French-LSF corpus, in our case the first task of the Rosetta-LSF corpus.
It consists in 194 French news items, each of 3 to 35 words in length, together with their LSF translations. For instance: ``\textit{L'Everest menacé de réchauffement climatique}'' (Everest threatened with global warming).
The translations were done by a deaf person selected for her experience in producing online LSF content on a regular basis. 

All of these news items have been encoded with AZee (section \ref{AZee}), resulting in a corpus of 194 AZee discourse expressions. 
These expressions reflect both the forms observed in the news in LSF and the meaning interpreted from it. 

From there, we were able to create a bank of AZee--text alignments which are correspondences between an AZee node and a text segment in French. 
In an AZee discourse expression, the root node necessarily covers the whole discourse in French, which already serves as an alignment. Besides, each node of the expression represents a portion of the news, which if a text segment is found to be equivalent in meaning, creates a new alignment of finer granularity. Considering each node of the AZee expressions allows to create entries on various levels (from whole news items to single words).
The alignments were created manually, guided by the three following principles. 
\begin{description}
    \item[Uniqueness] The same text segment cannot be aligned with different AZee sub-expressions.
    \item[Maximisation] The text segment has to correspond to the largest AZee sub-expression possible.
    \item[Objectivity] Only a text segment present in the news can be aligned: we want to be faithful to the example, not to account for generality by attempting to escape its scope.
\end{description}

After manually proceeding this way on all of the corpus, we found 1812 AZee--text alignments, collected in a file and encoded with the name of the text file in which the news in French is found, the first and last characters of the aligned French segment, the file name of the AZee discourse expression and finally the line number of the aligned AZee expression or sub-expression (node). For example: \[\textrm{RO1}\_\textrm{X0007.Titre1}\  10\  4\  \textrm{RO1}\_\textrm{X0007.Titre1.az}\  7\]

\section{Implementation}

\subsection{General algorithm} \label{general-algorithm}
Let $tr$ be the function that associates to a text query $q$
a set of possible translations for $q$ by analogy based on a corpus of aligned examples.
If the corpus contains alignments in which the text segment is exactly $q$, the set formed
by their aligned AZee expressions specifies an acceptable result for $tr(q)$.

Otherwise, as explained in §\ref{substitutions}, we consider the alignments whose text segments are ``close'' to $q$,
whose differences to $q$ are the ``anti-matches'',
whose translated counterparts in the aligned expression we hope to replace.
By doing this, the global structure of the aligned expression is kept to serve as a template for $tr(q)$,
in which substitutions are made.

Formally, for a given alignment between a text segment $txt$ and an expression $az$
where $txt$ qualifies as close to $q$, let:
\begin{itemize}
\item $\bar{m}_{1},...,\bar{m}_{N}$ be the anti-matches of $txt$, i.e.\ the parts in $txt$ that differ to $q$, where usually $N\le2$;
\item $c_{i}$ be the correction of $\bar{m}_{i}$ ($i\in1..N$), i.e.\ the wanted part of $q$ missing in $txt$;
\item $\bar{az}_{i}$ be the node in $az$ at the root of the sub-expression which translates $\bar{m}_{i}$ ($i\in1..N$).
\end{itemize}
With these notations, our approach is then to find a node $\bar{az}_{i}$ for each $i\in1..N$, and replace it with a translation of $c_{i}$.

For example, to translate query ``\textit{la présidente parle nerveusement}'', we could consider the text segment of the alignment below as close. In the AZee expression, rule \texttt{side-info}, with arguments \emph{focus} and \emph{info}, carries the meaning of \emph{focus} with additional (non-focused) information \emph{info} about it.

\begin{quote}
 \begin{description}
  \item[Text] ``\textit{le ministre de l'écologie parle nerveusement}''
  \item[AZee]
\begin{minipage}[t]{1\columnwidth}
\begin{verbatim}
:info-about
  'topic
  :side-info        (*)
    'focus
    :ministre
    'info
    :environnement
  'info
  :nerveusement
    'sig
    :parler
\end{verbatim}
\end{minipage}
 \end{description}
\end{quote}

The unique anti-match $\bar{m}_{1}$ is ``\textit{le ministre de l'écologie}'', and its correction $\bar{c}_1$ is ``\textit{la présidente}''. We would then want to identify the sub-expression marked \texttt{(*)} as its translation node $\bar{az}_{1}$, for which to substitute a translation of ``\textit{la présidente}''.

If no or several candidates for an $\bar{az}_{i}$ are found, it becomes a lot
less trivial to know what to substitute in $az$ regarding the $i$th anti-match.
For now, we implement translation failure in these cases, forcing each $\bar{az}_{i}$
to be found unique. The translations of $c_i$ can however be numerous, each one becoming an option for the $\bar{az}_{i}$ substitution.

Using our formal notations:
\begin{itemize}
\item finding $\bar{az}_{i}$ means finding a unique node $n$ such that $n$ is an acceptable translation for $\bar{m}_{i}$, in other words such that $n\in tr(\bar{m}_{i})$ ;
\item finding translations for $c_{i}$ implies simply to consider $tr(c_{i})$.
\end{itemize}
Then, any combination of $N$ substitutions $\bar{az}_{i}\rightarrow x,x\in tr(c_{i})$ can be applied to $az$ to create a translation of $q$. The set of all of them is therefore a subset of $tr(q)$ associated with the $txt$--$az$ alignment.
The full set can be specified as the union of such sets, iterating over all
known alignments with a text segment close to $q$.

The approach above yields a recursive definition of $tr(q)$ as it requires
values for $tr(\bar{m}_{i})$ and $tr(c_{i})$ for each anti-match encountered.
The base case for this recursion are the exact matches.
Besides, each anti-match is always a shorter segment than the initial query,
so the only condition for termination of this algorithm is to ensure that
corrections $c_{i}$ are always also shorter than the query, which is clearly
the typical case so adding this constraint will likely result in zero loss. 

More than termination, the problem is that of translation failure, which
is all the more likely to happen as the corpus of example alignments is small.
In such cases, we resort to a last fallback where we break the query down into
a partition of smaller text chunks, which we will translate separately and
concatenate in the result with the only reason that it follows the French order.
To do this we apply the AZee production rule \texttt{sign-supported-spoken} which
allows to build utterances based a spoken language literal sequence of items.

For example, one can chunk the query above into ``\textit{la présidente parle}'' + ``\textit{nerveusement}'', find a translation for each chunk separately, say (a) and (b) below, and propose (c) as a ﬁnal translation.

(a)~\begin{minipage}[t]{.2\textwidth}
\begin{verbatim}
:info-about
  'topic
  :président
  'info
  :parler
\end{verbatim}
\end{minipage}
~
(b)~\begin{minipage}[t]{.2\textwidth}
\begin{verbatim}
:nerveux
\end{verbatim}
\end{minipage}
~
(c)~\begin{minipage}[t]{.3\textwidth}
\begin{verbatim}
:sign-supported-spoken
  'units
  list
    :info-about
      'topic
      :président
      'info
      :parler
    :nerveux
\end{verbatim}
\end{minipage}
\medskip

For a given partition $\left\langle p_{1},p_{2},...,p_{n}\right\rangle $ of $q$,
the combinations $sign-supported-spoken(units=\left\langle x_{1},x_{2},...,x_{n}\right\rangle )$
with $x_{i}\in tr(p_{i})$ constitute a set of possible translations of $q$ with this technique.
By iterating on different partitions and joining all such sets, we generate a last, fallback specification of $tr(q)$.
This is also a recursive definition, whose recursive calls are applied to chunks ($p_{i}$) shorter
then $q$ by construction, so termination is guaranteed as well.

This fallback strategy produces poorer quality Sign Language, and indeed equivalent to literal (word-to-word)
translation if used systematically. But it does allow to juxtapose coarser-grain chunks of Sign when translation
succeeds without resorting to partitioning.

For example, the use of the rule \texttt{nerveux}, generating an additional manual sign meaning ``nervous'', can be judged as poorer LSF than that of \texttt{nerveusement} used further up, which generates a preferred and sufficient facial expression conveying the same meaning.
However, the first chunk was translated as a whole (using \texttt{info-about}), which did avoid the even poorer literal sign sequence below.

\begin{minipage}[t]{.7\columnwidth}
\begin{verbatim}
:sign-supported-spoken
  'units
  list
    :président
    :parler
    :nerveux
\end{verbatim}
\end{minipage}
\medskip

\subsection{Auxiliary text processing modules}

The practical implementation of the algorithm relies on several text processing modules enhancing analysis to find best correspondences in the existing corpus.

To allow for matching, antimatching and partitions, word-level tokenization is first performed by OpenNMT Tokenizer\footnote{\url{https://github.com/OpenNMT/Tokenizer}}, and flexibility is allowed when finding matching segments for punctuation and articles.

Then the core challenge is to define what kind of ``similarity'' in the source language can produce best candidates for target language generation. As can be seen in the example above, both semantics and syntax come into play to determine similar elements to be replaced or translated separately. In practice, we rely on two types of text analysis at different steps of the algorithm.

\begin{table*}
    \centering
    \begin{tabular}{llll}
    \hline Alignment text & \pbox{1cm}{Common tokens} & Length & Ratio \\
    \hline \pbox{20cm}{le superéthanol n'est proposé que dans 1 000 stations-service en \\ france , comme ici \textbf{dans la}  \textbf{banlieue de} bordeaux .} & 4 & 22 & 0.18 \\
    \hline \pbox{20cm}{comme ici \textbf{dans la banlieue de} bordeaux} & 4 & 7 & 0.57 \\
    \hline \textbf{la banlieue de} bordeaux & 3 & 4 & 0.75 \\
    \hline \pbox{20cm}{situé \textbf{dans la} province du guizhou , en chine , le mont fanjing attire \textbf{de} \\ nombreux touristes venus découvrir la richesse de ce paysage montagneux .} & 3 & 26 & 0.12 \\
    \hline \pbox{20cm}{\textbf{la} villa noailles à hyères \textbf{dans} le var est un château cubiste construit dans \\ les années folles , à la demande d'un couple \textbf{de} mécènes avant-gardiste .} & 3 & 29 & 0.10 \\
    \hline 
    \end{tabular}
    \caption{Antimatchable alignments to translate ``dans la banlieue de Gerstheim" (\textit{in the suburbs of Gerstheim})}
    \label{tab:antimatches}
\end{table*}

To find the best anti-matches in the current database and replace them by corrections, we use string matching and consider as ``anti-matchable'' all alignments that have tokens in common with the candidate text.
Best matches have been empirically set as the ones with the maximum tokens in common, and either the minimum length in number of tokens or the best ratio of similar tokens over total tokens.
For example, to translate ``\textit{dans la banlieue de Gerstheim}'' (in the suburbs of Gerstheim) by anti-match, the alignments with most tokens in common found in the database are described in Table \ref{tab:antimatches}.
We see that selecting, among the alignments with the highest number of tokens in common (4), the alignment with the lowest length or best ratio between number of similar tokens and length enables to retrieve the most relevant alignment for anti-match: ``\textit{comme ici \textbf{dans la banlieue de} Bordeaux}'' (like here in the suburbs of Bordeaux).
Other sets of metrics could be used successfully, as we found that the selection and ranking of alignments for anti-matching strategy significantly affects the results of the algorithm.

When matching and anti-matching approaches fail, we resort to partitions determined by navigating the syntactic dependency tree obtained using spaCy\footnote{\url{https://spacy.io}}, open-source Python library with off-the-shelf pretrained models and optimized pipelines for Natural Language Processing. 
For instance, for the sentence ``\textit{Le couvre-feu cette semaine n'est pas encore arrêté}'' (curfew this week has not yet been stopped), we consider as candidate partitions:
\begin{itemize}
    \item ``\textit{le couvre-feu}'' / ``\textit{cette semaine n'est pas encore arrêté}'';
    \item ``\textit{le couvre-feu cette semaine n'est}'' / ``\textit{pas encore}'' / ``\textit{arrêté}'';
    \item ``\textit{le couvre-feu}'' / ``\textit{cette semaine}'' / ``\textit{n'est pas encore arrêté}''.
\end{itemize}

Our tree exploration makes that some possible partitions such as the following are not explored: ``\textit{Le couvre-feu}'' / ``\textit{cette semaine}'' / ``\textit{n'est pas}'' / ``\textit{encore}'' / ``\textit{arrêté}''.
Dependency parsing does not explore all possible partitions of a sentence but at least constrains the exploration to syntactically valid chunks.

\section{Test}

\subsection{Test set}
Admittedly, the data base is still too small for us to claim a system up to any usable scale. So to test our system, we decided to build a test set by creating sentences mixing segments from different entries of our corpus, and evaluate the produced outputs.

Our test set is composed of 15 sentences and allows to test the algorithm, as presented in the next section. 
For instance, the sentence ``\textit{Recul de l'âge légal à la retraite : c'est ce que proposent des retraités pour leurs enfants}'' (Increase of the retirement age: pensioners propose it for their children) was added to the test set and created from the following sentences of the corpus:
\begin{itemize}
    \item \textit{\textbf{Recul de l'âge légal à la retraite :} ``Il ne faut pas prendre les Français pour des canards sauvages'', lance Valérie Pécresse.} (Increase of the retirement age: ``We should not take the French for a ride'', shouts Valérie Pécresse.)
    \item \textit{Des routes nationales bientôt privatisées ? \textbf{C'est ce que proposent} les sociétés d'autoroutes dans une note interne.} (National roads soon to be privatised? Motorway companies propose it in an internal memo.)
    \item \textit{Solidarité : une ancienne abbaye accueille \textbf{des retraités}} (Solidarity: a former abbey hosts pensioners.)
    \item \textit{Au Japon, des dizaines de pères français se battent désespérément \textbf{pour} voir \textbf{leurs enfants.}} (In Japan, dozens of French fathers are desperately fighting to see their children.)
\end{itemize}

\subsection{Algorithm on an example}
This section describes the steps taken by the algorithm run on the following example taken from the test set, and the produced AZee description results.
\begin{quote}
``\textit{Alsace : de grands chefs ont vendu leur vaisselle pour les plus modestes comme ici dans la banlieue de Gerstheim.}'' (Alsace: great chefs sold their crockery for the poor like here in the suburbs of Gerstheim.)
\end{quote}

The whole sentence is tested, first for exact matches, then for anti-matching segments but to no avail. So it falls back to partitioning the query, breaking it down into 3 smaller segments as follows: ``\textit{Alsace}'' / ``\textit{de grands chefs}'' / ``\textit{ont vendu leur vaisselle pour les plus modestes comme ici dans la banlieue de Gerstheim}''.

Each segment above is then used as a new (simpler) input query in a recursive call to the algorithm, reported below. See fig.~\ref{fig:partial-matches} for the referenced AZee expression matches.
\begin{description}
 \item[``\textit{Alsace}''] An exact-match (d) is found, which is directly returned as an acceptable translation for this segment.
 \item[``\textit{de grands chefs}''] Similarly, an exact-match (e) is found.
 \item[``\textit{ont vendu leur vaisselle pour} ...''] There is no exact match, and no anti-matching segment is found either to translate this text chunk. So again, the query is broken down into the smaller sub-queries below.
 \begin{description}
  \item[``\textit{ont vendu leur vaisselle}''] Exact match (f1) found.
  \item[``\textit{pour les plus modestes}''] Exact match (f2) found.
  \item[``\textit{comme ici dans la banlieue de Gerstheim}''] No exact match, but similar segment found, aligned with (f3'): ``\textit{comme ici dans la banlieue de Bordeaux}'', anti-match ``\textit{Bordeaux}'' to be corrected with ``\textit{Gerstheim}''.
  Both the anti-match and the correction trigger a recursive call, the former to decide on a node to change in (f3'); the latter to find what to change it for.
  \begin{description}
   \item[``\textit{Bordeaux}''] Exact match \texttt{:Bordeaux} found in the alignment base, recognised and unique in (f3')---marked \texttt{(**)} in fig.~\ref{fig:partial-matches}.
   \item[``\textit{Gerstheim}''] Exact match (f3'') found in the alignment base.
  \end{description}
  Let (f3) be the expression (f3') with (f3'') instead of node \texttt{(**)}; (f3) is a resulting translation for this query.
 \end{description}
 Now that each segment of the inner partition has found a translation, a result can be produced by creating a \texttt{sign-supported-spoken} expression with units (f1), (f2) and (f3) in this order.
\end{description}
Finally and in the same way, a result can be proposed for the outer partition using \texttt{sign-supported-spoken}. The overall expression is therefore the following:

\begin{minipage}[t]{.7\columnwidth}
\begin{verbatim}
:sign-supported-spoken
  'units
  list
    (d)
    (e)
    :sign-supported-spoken
      'units
      list
        (f1)
        (f2)
        (f3)
\end{verbatim}
\end{minipage}
\medskip

\begin{figure*}
\begin{center}
(d)~\begin{minipage}[t]{.25\textwidth}
\begin{verbatim}
:category
  'cat
  :info-about
     'topic
     :Est
     'info
     :info-about
        'topic
        :France
        'info
        :zone
  'elt
  :info-about
     'topic
     :appartenance
     'info
     :Alsace
\end{verbatim}
\end{minipage}
~
(e)~\begin{minipage}[t]{.25\textwidth}
\begin{verbatim}
:category
  'cat
  :side-info
     'focus
     :multiplicity
        'elt
        :une personne
     'info
     :zone
  'elt
  :chef cuisinier
\end{verbatim}
\end{minipage}
~
(f1)~\begin{minipage}[t]{.3\textwidth}
\begin{verbatim}
:info-about
  'topic
  :là
  'info
  :info-about
    'topic
    :all-of
      'items
      list
        :assiette
        :assiette
    'info
    :multiplicity
       'elt
       :vendre
\end{verbatim}
\end{minipage}
\end{center}
\vspace{1cm}
\begin{center}
(f2)~\begin{minipage}[t]{.25\textwidth}
\begin{verbatim}
:info-about
  'topic
  :pour
  'info
  :side-info
    'focus
    :multiplicity
      'elt
      :une personne
    'info
    :info-about
       'topic
       :comment dire
       'info
       :difficile
\end{verbatim}
\end{minipage}
~
(f3')~\begin{minipage}[t]{.25\textwidth}
\begin{verbatim}
:info-about
  'topic
  :exemple
  'info
  :info-about
    'topic
    :aussi
    'info
    :info-about
      'topic
      :ici
      'info
      :info-about
        'topic
        :side-info
          'focus
          :Bordeaux  (**)
          'info
          :banlieue
        'info
        :là
\end{verbatim}
\end{minipage}
~
(f3'')~\begin{minipage}[t]{.3\textwidth}
\begin{verbatim}
:category
  'cat
  :ville
  'elt
  :fingerspelling
    'letters
    list
      .G
      .E
      .R
      .S
      .T
      .H
      .E
      .I
      .M
\end{verbatim}
\end{minipage}
\end{center}
\begin{quote}
In English:
\begin{description}
    \item[(d)] The Alsace region in the East of France
    \item[(e)] Chefs
    \item[(f1)] Sell crockery
    \item[(f2)] For the poor (people)
    \item[(f3')] Like here in the suburbs of Bordeaux
    \item[(f3'')] Gerstheim
\end{description}
\end{quote}
\caption{Aligned AZee expressions matched in algorithm run}
\label{fig:partial-matches}
\end{figure*}

\section{Discussion}

First, our anti-matches approach has some advantages, compared to a sequence-based one. 
Indeed, structures that are specific to LSF can be found in the final translations, which is not the case when the language is reduced to a sequence of glosses.

In addition, the approach produces results with a certain form of creativity. In LSF, paraphrases or additions are commonly used, and indeed part of our corpus as initially delivered by the translator at the time of video corpus creation. These elements were later aligned as examples, thus frequently appear in the generated translations, although not always strictly necessary. See the example for ``\textit{Alsace}'', which although a single sign exists, is translated to the whole expression (d), typical of LSF when no context yet exists.

Moreover, the output of the algorithm is a set of translations
(built from the various substitution combinations), not necessarily a single expression. This in a way accounts for the reality of the translation task.
For example, to translate ``\textit{Emmanuel Macron}'' into LSF, different possibilities have been used by the translator, hence the different possible AZee output expressions (g), (h) and (i) below.

(g)~\begin{minipage}[t]{.5\columnwidth}
\begin{verbatim}
:Emmanuel Macron
\end{verbatim}
\end{minipage}

(h)~\begin{minipage}[t]{.5\columnwidth}
\begin{verbatim}
:side-info
  'topic
  :Emmanuel Macron
  'info
  :président 
\end{verbatim}
\end{minipage}

(i)~\begin{minipage}[t]{.5\columnwidth}
\begin{verbatim}
:category
  'cat
  :side-info
    'topic
    :une personne
    'info
    :président
  'elt
  :Emmanuel Macron 
\end{verbatim}
\end{minipage}
\medskip

In our test set, the number of translations proposed for a query ranges from 1 to 12 (average: 4).
At the moment, the order in which the AZee expressions are output is irrelevant. One prospect for this algorithm is to rank them according to some heuristics, for example constraints on preferred AZee rule combinations. 

The work presented also has limitations. 
We can observe that some anti-matches are incorrect, for instance: ``\textit{des personnes pro-Brexit}'' (pro-Brexit people) vs ``\textit{des personnes manifestent}'' (people demonstrate). The syntactic categories of the anti-match and its correction are not the same (adjective vs. verb), which creates problems during the translation process. If we want to translate ``\textit{des personnes pro-Brexit sont dans la rue}'' (pro-Brexit people are in the street), the algorithm suggests ``\textit{pro-Brexit}'' as an anti-match for ``\textit{manifestent}'', but the result is syntactically unacceptable: ``\textit{*des personnes manifestent sont dans la rue}''.
The syntactic category of each phrase should be taken into account to prevent such errors and to improve the anti-matching results.

Finally, a considerable number of fallbacks are present in the output of the algorithm: 3 per result on average.
As explained in section~\ref{general-algorithm}, this is not ideal, and the size of our corpus is undoubtedly a contributing factor: if we increase the number of examples and alignments, the number of fallbacks will decrease and the quality of the translations should hopefully improve.

\section{Conclusion and prospects}

We have presented a new system of automatic translation from text to AZee, based on an example-based machine translation approach, the hierarchical representation of SL AZee and an aligned corpus of French text and AZee descriptions extracted from the Rosetta-LSF corpus. A prototyping implementation of the system has been made and tested on some examples, thus providing a proof of concept.

The capacities of this system and the size of the corpus still need to be extended before real evaluations can be carried out. But we can already stress that the evaluation of such a system will not be easy, since it proposes a translation from one language to a representation of another language, not readable directly.

Metrics for evaluating the quality of translations, such as the one proposed in the European QT21 project\footnote{\url{https://www.qt21.eu}}, provide a scoring sheet with types of errors produced by the translation system, which allows to highlight the shortcomings of the systems and the aspects to improve.
This project has proposed Multidimensional Quality Metrics (MQM), a framework for describing and defining custom translation quality metrics. It provides a flexible vocabulary of quality issue types, a mechanism to generate quality scores, and mappings to other metrics.

Some of the error categories are defined assuming text as a target, which does not apply in our case. A category called ``fluency'' allows us to evaluate the quality of an utterance, regardless of whether it is the result of a translation. In our case, the target is not even a language utterance, thus this category will need some adjustments.
What remains is the category of errors linked to the translation process itself, categorised as ``accuracy''.
Accuracy errors happen when the target text does not reflect the source text. This could be by addition (of content not present in the source), mistranslation (output content not accurately aligned with the source) or omission (of content that was present in the source).

It would be interesting to study if this kind of evaluation could be adapted to our system.
The issue is to define these categories in the case of SL. It is common or indeed often preferred in SL to introduce contextual information, for example expression (d) figure~\ref{fig:partial-matches} for ``Alsace'', which should not be judged as an unwanted addition.

Furthermore, as AZee can be used to generate virtual signer animations which are directly ``readable'' by language users, fluency error categories could be taken into consideration after this step to complete the evaluation. The establishment of a robust and comprehensive evaluation protocol is clearly a subject of study in its own right, which needs to be pursued in the near future.

\section*{Acknowledgements}

This work has been funded by the Bpifrance investment project ``Grands défis du numérique'', as part of the ROSETTA project (RObot for Subtitling and intElligent adapTed TranslAtion).

We thank Noémie Churlet, Raphaël Bouton and Media'Pi! for their commitment to this project, which would not have had the same validity and impact without them.

\bibliographystyle{plain}
\bibliography{artick-trad-arxiv-2022}

\end{document}